%% file: main.tex
\documentclass[letterpaper, 10pt, conference]{ieeeconf}      

\IEEEoverridecommandlockouts                              

\usepackage{graphics} 
\usepackage{epsfig} 
\usepackage{amsmath} 
\usepackage{amssymb}  
\usepackage{bm} 

\input{macros.tex}  
\usepackage{siunitx} 
\usepackage{textcomp, gensymb}
\usepackage{gensymb} 

\usepackage{pgfplots}
\pgfplotsset{compat=newest}
\usetikzlibrary{plotmarks}
\usetikzlibrary{arrows.meta}
\usepgfplotslibrary{patchplots}
\usepackage{grffile}
\pgfplotsset{plot coordinates/math parser=false}
\newlength\fwidth
\newlength\fheight
\usepackage{url}

\usepackage{cite}

\usepackage{flushend} 
\usepackage{subcaption}
\usepackage[caption=false,font=small,labelformat=simple]{subfig} 

\title{\LARGE \bf
Design and Development of Modular Limbs for Reconfigurable Robots on the Moon}

\author{Gustavo H. Diaz$^{1}$, A. Sejal Jain$^{1}$, Matteo Brugnera$^{1}$, Elian Neppel$^{1}$, \\Shreya Santra$^{1}$, Kentaro Uno$^{1}$ and Kazuya Yoshida$^1$
    \thanks{$^*$This work was supported by JST Moonshot R\&D Program, Grant Number JPMJMS223B.}
    \thanks{
    $^1$All authors are with the Space Robotics Lab. (SRL) in the Department of Aerospace Engineering, Graduate School of Engineering, Tohoku University, Sendai 980--8579, Japan. 
    }%
    \thanks{\textit{The corresponding author is Gustavo H. Diaz.}}%
    \thanks{(E-mail: \tt{gustavo.diaz@dc.tohoku.ac.jp})}%
}%

\begin{document}

\maketitle
\thispagestyle{empty}
\pagestyle{empty}

\begin{abstract}
In this paper, we present the development of 4-DOF robot limbs, which we call Moonbots, designed to connect in various configurations with each other and wheel modules, enabling adaptation to different environments and tasks. These modular components are intended primarily for robotic systems in space exploration and construction on the Moon in our Moonshot project. Such modular robots add flexibility and versatility for space missions where resources are constrained.
Each module is driven by a common actuator characterized by a high torque-to-speed ratio, supporting both precise control and dynamic motion when required. This unified actuator design simplifies development and maintenance across the different module types. The paper describes the hardware implementation, the mechanical design of the modules, and the overall software architecture used to control and coordinate them. Additionally, we evaluate the control performance of the actuator under various load conditions to characterize its suitability for modular robot applications.
To demonstrate the adaptability of the system, we introduce nine functional configurations assembled from the same set of modules: 4DOF-limb, 8DOF-limb, \textit{vehicle, dragon, minimal, quadruped, cargo, cargo-minimal, and bike}. These configurations reflect different locomotion strategies and task-specific behaviors, offering a practical foundation for further research in reconfigurable robotic systems.
\end{abstract}

\section{Introduction}\label{introduction}
\begin{figure}[t]
  \centering
  \includegraphics[width=.95\linewidth]{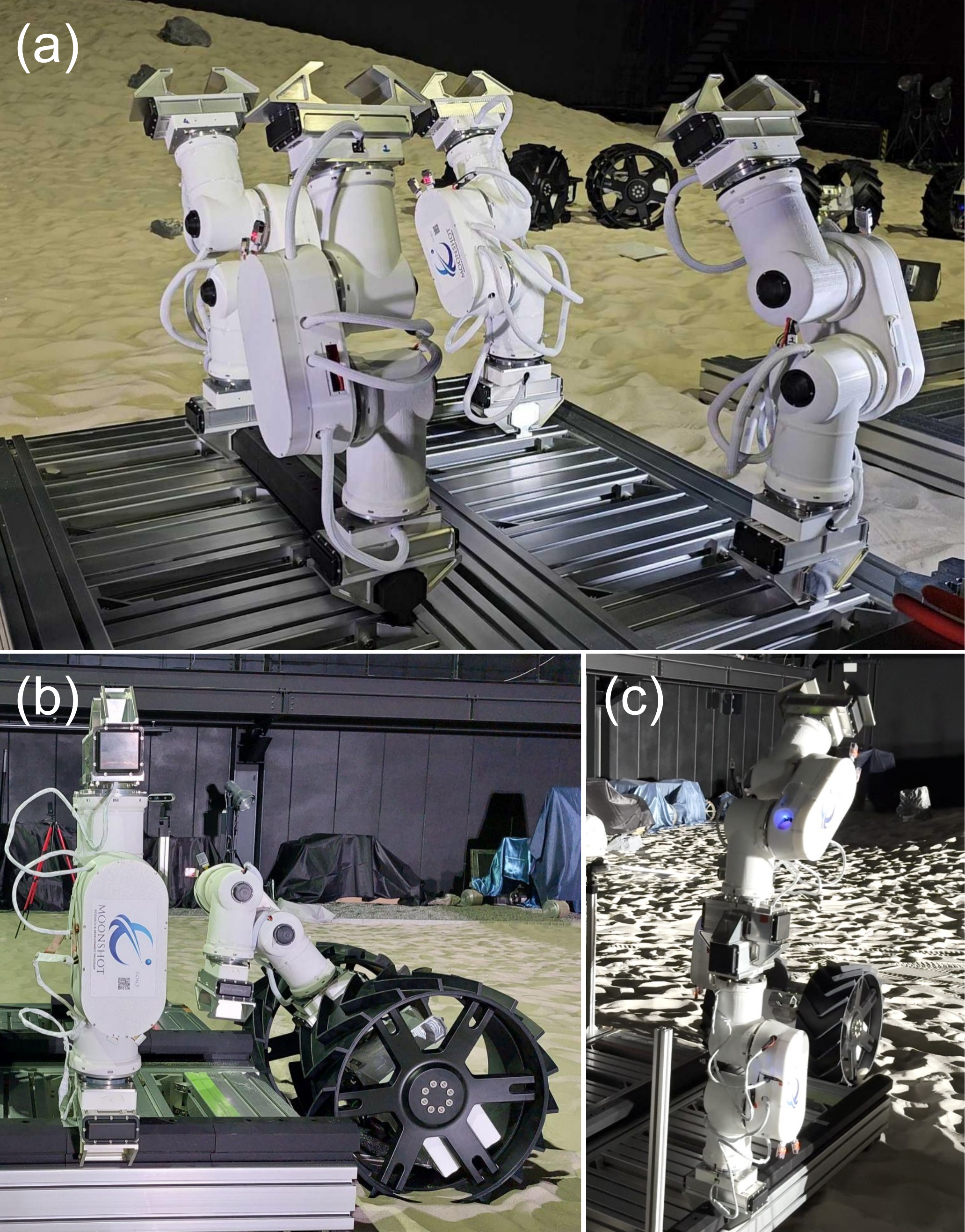}
  \caption{(a) Developed four degree-of-freedom (DOF) modular limbs arranged on a palette. (b) One of the limb grasping to the wheel module to achieve mobility. (c) 8-DOF robot manipulator formed by the mutual connection of the two 4-DOF limbs.}
  \label{4dof_limbs_at_jaxa}
  \vspace{-2mm}
\end{figure}
In the context of the Moonshot Project Goal~3~\cite{moonshot}, we are developing modular robots for self-assembly and construction tasks on the Moon~\cite{moonbot}. Our earlier work focused on the semi-autonomous assisted assembly of a four-legged robot using an \textit{xarm-7} manipulator~\cite{gustavo}. While this approach demonstrated the feasibility of modular assembly, it addressed a highly constrained scenario: all components were fixed on a table within a controlled environment, and the assembled robot had minimal payload capacity, limiting its capability for practical construction tasks. 
To overcome these limitations, we focus on creating a more versatile modular robotic platform capable of carrying substantial payloads and supporting multiple configurations for mobility, manipulation, and mobile manipulation. The core building blocks of this system are the \textit{limb modules}, which serve as the connecting elements for diverse robot configurations. In our previous design~\cite{moonbot0}, these modules provided 3~degrees of freedom (DOF). In this work, we introduce a new 4-DOF minimal module design that increases versatility while keeping hardware complexity low.
The proposed limb modules can operate independently as robot limbs or as a steering/suspension \textit{bridge} connecting wheel modules, enabling the creation of mobile platforms suitable for further integration into complex robotic systems, including mobile manipulators. They also provide sufficient DOFs to act as legs in quadruped configurations or as arms for manipulation. The DOF arrangement follows a symmetric roll–pitch–pitch–roll configuration, driven by a custom-designed actuator that meets the torque, speed, and control requirements. A robust gripper mechanism ensures stable and repeatable inter-module connections. We developed four such limbs and tested them in field experiments at JAXA’s Space Exploration Test Field in Sagamihara campus~\cite{Tansa_X}, as shown in Fig.~\ref{4dof_limbs_at_jaxa}. The contributions of this work are as follows:  
\begin{itemize}
    \item Design and implementation of a compact, high-performance actuator capable of delivering over \SI{75}{Nm} torque and achieving a maximum speed of \SI{26}{rpm}, suitable for both manipulation and locomotion;  
    \item Development of a macro-modular, heterogeneous robotic architecture composed of 4-DOF limb modules and wheel modules, enabling diverse configurations for various tasks and environments;  
    \item Integration and experimental validation of these modules in multiple configurations, including field testing in unstructured terrain.
\end{itemize}


\section{Related Work}\label{related_work}

As modular reconfigurable robotic systems have evolved, the need for systematic classification frameworks has become evident to analyze diverse designs, evaluate capabilities, and guide future developments. Chennareddy \textit{et al.}~\cite{chennareddy} proposed a comprehensive scheme that organizes modular robots into four primary dimensions: structure, reconfiguration method, form factor, and locomotion, each with multiple subcategories.
According to this framework, the modules developed in this work are classified as follows. \textbf{Structure:} lattice-like, with actuators in fixed positions; multiple limbs can be linked to form chain-like manipulators or combined with wheels for other morphologies. \textbf{Reconfiguration:} Deterministic, via teleoperation, joint-level control, and inverse kinematics (IK) commands. \textbf{Form factor:} macro-scale modules. \textbf{Locomotion:} Coordinated, achieved by coupling limbs to wheel modules or central body; implemented in configurations such as the dragon, cargo, and quadruped modes. 

A second classification approach is provided by the taxonomy in~\cite{survey_framework}, which characterizes Modular and Reconfigurable Robotic (MRR) systems along three principal axes: connector type, actuation method, and system-level homogeneity. Using this scheme, the proposed limb modules are described as follows. \textbf{Connector type:} monogamous, with each connector forming a single link at a time. For limb-to-limb connections, the interface is genderless, enabling symmetric, bidirectional coupling. When connecting to modules with predefined mating fixtures, such as wheel modules, the connector acts as male, actively engaging in insertion. The gripper mechanism used for this purpose is shown in Fig.~\ref{limb_components}. \textbf{Actuation method:} joint-based actuation with high accuracy and minimal backlash. \textbf{Homogeneity:} the limbs are identical and can operate in a homogeneous configuration; when combined with wheel modules, the system becomes heterogeneous. This qualifies the architecture as functionally reconfigurable, allowing task-level adaptation through module rearrangement rather than extensive structural redesign.

Several existing modular robotic systems share partial similarities with our approach. THU-QUAD II~\cite{multi-task-control} is a quadruped robot capable of switching between mammal-type and sprawling-type postures to traverse varied terrain, including stairs, doorsills, and uneven surfaces. This is achieved via extended joint ranges (up to $330^{\circ}$) and a control framework mapping environmental features to gait and stance selection. In contrast, the system presented here uses modularity-driven reconfigurability: each limb can function as a leg, arm, or structural link, enabling transformation across quadruped, wheeled, manipulator, and hybrid configurations. In quadruped mode, our limbs can replicate the postures described in~\cite{multi-task-control}, but can also reconfigure into a wheeled vehicle for efficient travel over flat lunar regolith, or into a \emph{Dragon} configuration -- two limbs and two wheels are serially connected form -- for manipulation tasks.

In~\cite{cobot,snapbot}, a plug-and-work reconfigurable collaborative robot (Cobot and Snapbot) is introduced, with interchangeable elbow joints, link modules, and end-effectors. While this system supports plug-and-play assembly, demonstrations are limited to static arm configurations (3--5-DOF) and do not address locomotion or heterogeneous assemblies.
The Snake Monster robot~\cite{snake} also emphasizes reconfigurability, distributed actuation, and terrain-adaptive locomotion. Our modular limb architecture extends these principles by supporting both locomotion and manipulation within the same heterogeneous system, enabling broader functional adaptability.

\section{Modular Limb Design and Development}\label{design_development}
\begin{figure}[t]
  \centering
  \includegraphics[width=\linewidth]{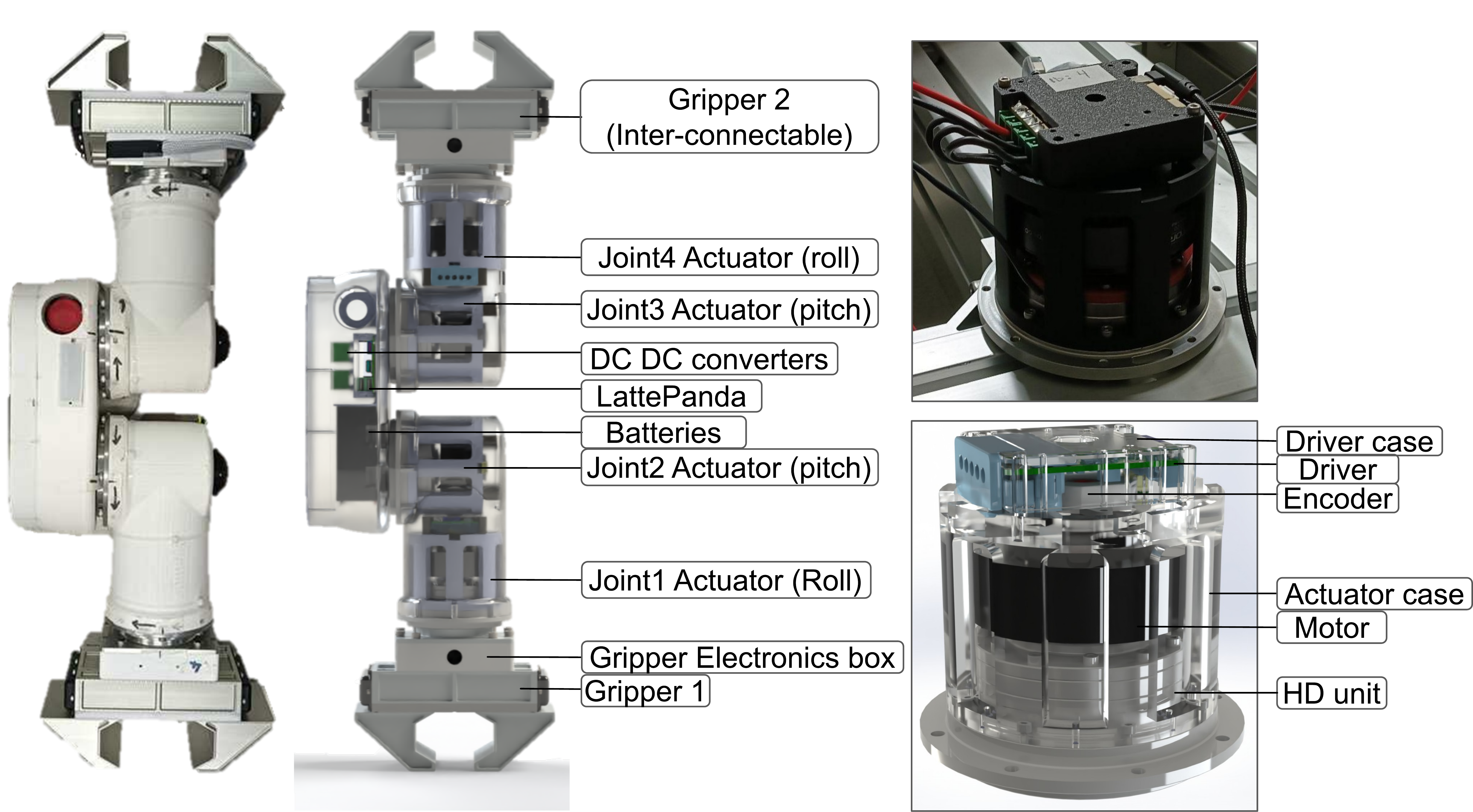}
  \caption{Developed robotic limb and actuator internal components.}
  \vspace{-2mm}
  \label{limb_components}
\end{figure}

This section describes the core components, hardware architecture, and software framework of the limb modules. An overview of the main components is provided in \fig{limb_components}. Each module integrates custom-developed actuators for the joints, a gripper, a microprocessor, batteries, and DC-DC converters. The actuators are detailed in \ref{subsec:actuator_design}, while the grippers were first presented in \cite{moonbot}. 

\subsection{Actuator Design}\label{subsec:actuator_design} 
Each actuator comprises four primary components: a strain-wave speed reducer, a brushless DC (BLDC) outrunner motor, a motor driver, and an encoder. Outrunner motors, typically used in UAVs, were chosen for its low weight, high torque at low rotational speeds, and rapid acceleration/deceleration capabilities. A Harmonic Drive \textit{CSD-25-160-2A-GR} was employed for its high precision and torque density. A summary of the actuator specifications is given in \tab{tab:actuator_specs}.

\begin{table}[tb]
\scriptsize
\caption{\label{tab:actuator_specs} Specifications of the actuator components.}
\vspace{-2mm}
\begin{tabular}{|l|l|l|}
\hline
{\bf Component}                                               & {\bf Type, Manufacturer}                                                                         & {\bf Specifications}                                                                                                                                             \\ \hline
\begin{tabular}[c]{@{}l@{}}Speed\\ Reducer\end{tabular} & \begin{tabular}[c]{@{}l@{}}Strain wave gear,\\ Harmonic Drive (HD)\end{tabular}           & \begin{tabular}[c]{@{}l@{}}Reduction ratio: 1:160\\ Torque (Rated/Peak): 47 Nm/123 Nm\end{tabular}                                                  \\ \hline
\begin{tabular}[c]{@{}l@{}}HD\\ Housing\end{tabular}    & Custom, HERO Lab.                                                                           & Duralumin                                                                                                                                                         \\ \hline
Motor                                                   & \begin{tabular}[c]{@{}l@{}}BLDC Outrunner,\\ T-Motor\end{tabular}                         & \begin{tabular}[c]{@{}l@{}}Max. Power: 3180 W\\ Peak torque: 5.8 Nm\end{tabular}                                                                           \\ \hline
\begin{tabular}[c]{@{}l@{}}Motor\\ Driver\end{tabular}  & \begin{tabular}[c]{@{}l@{}}Field-Oriented\\ Control (FOC),\\ Odrive Robotics\end{tabular} & \begin{tabular}[c]{@{}l@{}}Max. Power: 5 kW\\ Peak current: 120A\\ Continous current:  70 A\\ Control modes: position,\\ velocity and current\end{tabular} \\ \hline
Encoder                                                 & \begin{tabular}[c]{@{}l@{}}Capacitive,\\ Same Sky\\ (CUI devices)\end{tabular}            & \begin{tabular}[c]{@{}l@{}}Model: AMT212C / AMT232B\\ Abs. res.: 12-bit multiturn/14-bit\\ Comm. Interface: RS485/SSI\end{tabular}                         \\ \hline
\end{tabular}
\end{table}

\subsection{Limb Design} 
\begin{figure}[t]
  \centering
  \includegraphics[width=.95\linewidth]{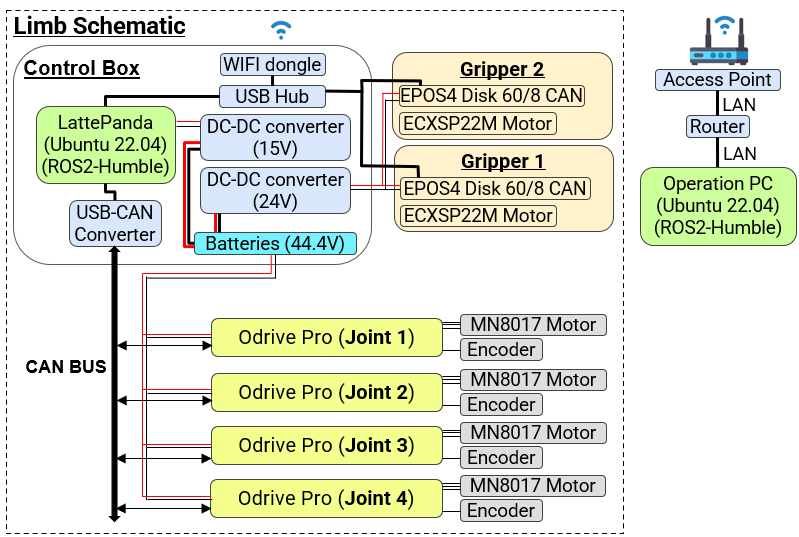}
  \caption{Hardware architecture of the limb module.}
  \vspace{-2mm}
  \label{limb_schematic}
\end{figure}

The limb comprises three main links and a gripper. Two of the links are mechanically identical, each containing two perpendicular actuators, while the third link functions as the control box. The modular design facilitates compact packaging, simplified manufacturing, and ease of assembly and maintenance. The control box houses a LattePanda 3 Delta single-board computer, two DC-DC converters, an emergency stop switch, and two 6S \SI{2400}{mAh} LiPo batteries connected in series. One converter powers the LattePanda, while the other supplies the grippers. The motor drivers are powered directly from the batteries, enabling the absorption of regenerated energy during active braking. All motor drivers are connected via a daisy-chained CAN bus, which interfaces with the LattePanda through a USB–CAN converter. Communication with the control PC is achieved over a local wireless network, where the LattePanda connects via a USB Wi-Fi dongle and the control PC is linked via LAN. The hardware architecture is shown in \fig{limb_schematic}.

\subsection{Software Architecture} 

\begin{figure}[t]
  \centering
  \includegraphics[width=0.85\linewidth]{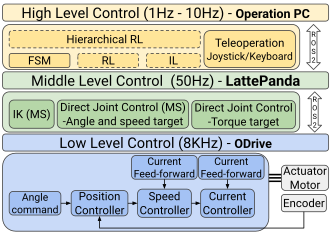}\vspace{-2mm}
  \caption{Three-layered structure of the limb control software.}
  \vspace{-2mm}
  \label{software_architecture}
\end{figure}

ROS~2 serves as the core middleware for control and communication. The software stack is organized into three hierarchical levels, illustrated in \fig{software_architecture}:
\begin{itemize}
    \item \textbf{Low level:} Interfaces directly with the ODrive motor drivers, which implement nested FOC controllers operating at \SI{8}{\kilo\hertz}. Communication is established through the official ROS~2 ODrive package\footnote{\url{https://github.com/odriverobotics/ros_odrive}(All URLs accessed: 2025-07-26)}, with custom nodes added for compatibility with standard \texttt{joint\_states} messages.
    \item \textbf{Middle level:} Provides joint command interfaces and inverse kinematics (IK) through either a standalone ROS~2 node (for calibration and testing) or the Motion-Stack (MS) framework~\cite{ms_api}, which supports synchronized multi-joint and IK-based control.
    \item \textbf{High level:} Currently supports joystick and keyboard teleoperation, with planned integration of reinforcement learning (RL) and other learning-based control strategies, which would increase adaptability and robustness to novel tasks and environments.
\end{itemize}

\begin{figure}[t]
  \centering
  \includegraphics[width=.9\linewidth]{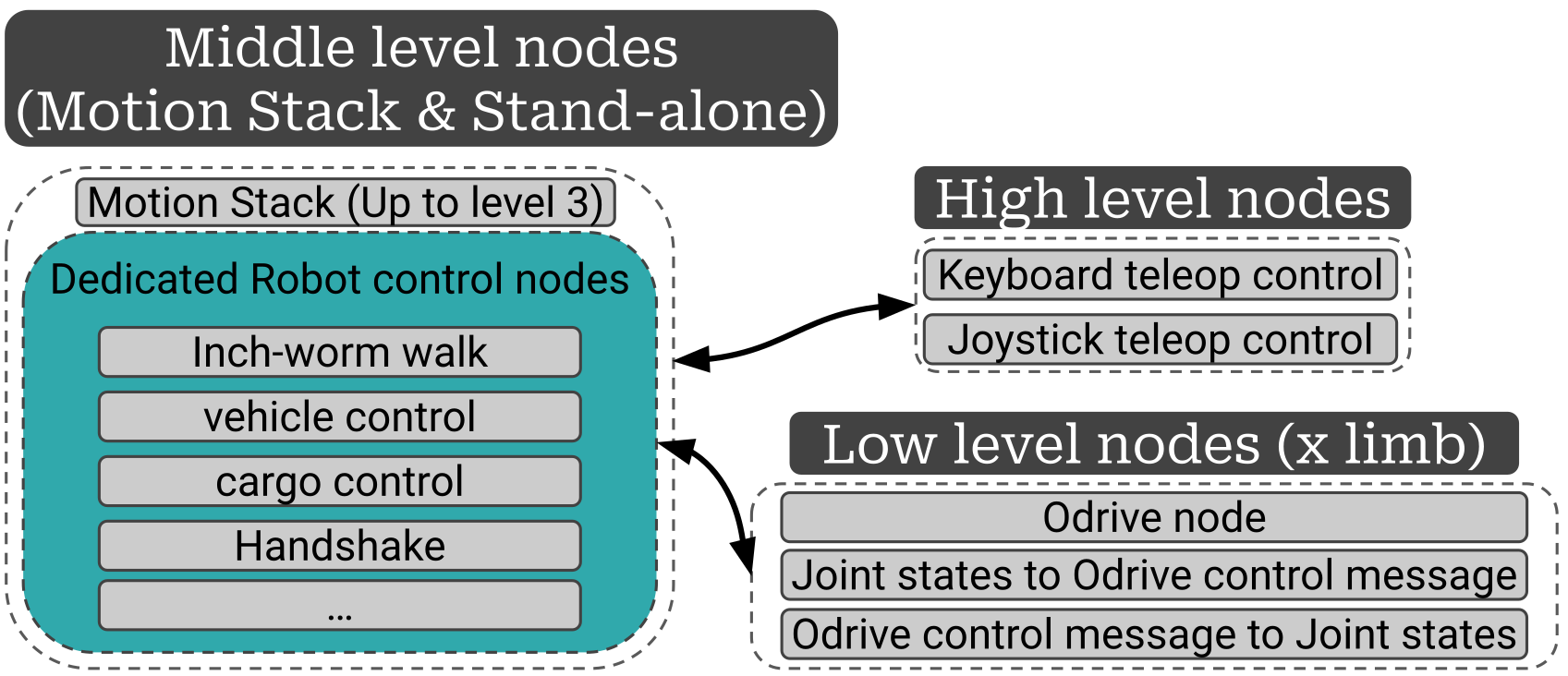}
  \caption{Implemented ROS~2 nodes for limb control.}
  \vspace{-2mm}
  \label{implemented_nodes}
\end{figure}

A more detailed view of the ROS~2 node implementation is presented in \fig{implemented_nodes}. At the low level, three nodes manage CAN communication with ODrives and interface with \texttt{joint\_states}. Middle-level nodes handle both standalone and MS-based control. High-level nodes manage teleoperation. Specific routines implemented in the middle level include \textit{inchworm motion} (pallet-top displacement via grasping fixtures), \textit{limb-to-limb connection}, vehicle mode control, and \textit{limb handshake}.

\section{Hardware Evaluation}\label{evaluation}

\subsection{Actuator}
The actuator modules were evaluated through static and dynamic load capacity tests, as well as position control step response experiments under different loads and control modes. The following subsections detail the experimental setup, procedures, and results.

\subsubsection{Static load test} 
\begin{figure}[t]
  \centering
  \begin{subfigure}[t]{1.0\linewidth}
    \includegraphics[width=\linewidth]{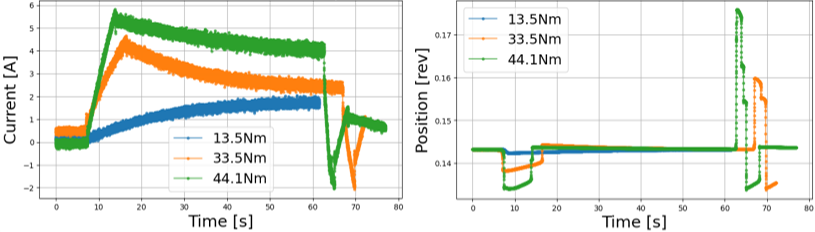}
    \caption{Actuator current (left) and position (right) over time for different applied loads.}
    \vspace{-2mm}
    \label{actuator_load_currents_and_pos}
  \end{subfigure}
  \begin{subfigure}[t]{1.0\linewidth}
    \vspace{2mm}
    \centering
    \includegraphics[width=.65\linewidth,clip]{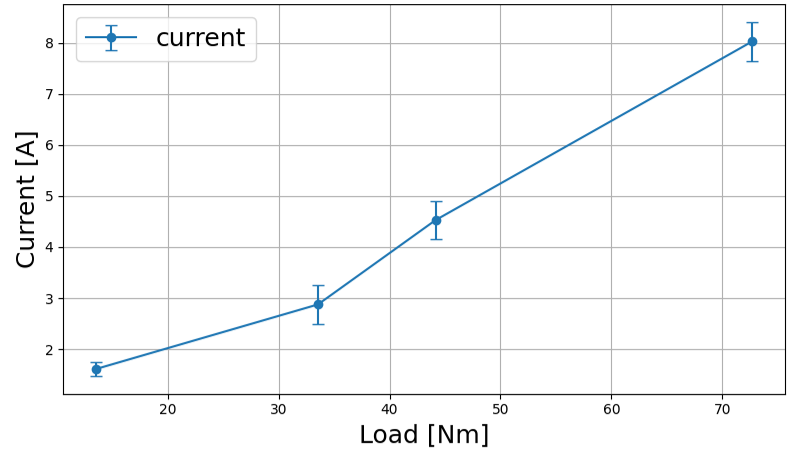}
    \caption{Average actuator current v.s. applied loads.}
    \label{actuator_current_vs_load}
  \end{subfigure}
  \caption{Static load test for the custom made joint actuator.}
  \label{static_load_test}
\end{figure}

The objective of this test was to determine the maximum payload capacity of the actuators when maintaining a fixed position. Loads were applied incrementally to the actuator output shaft by connecting a lever and hanging weights on it, and the resulting motor current and position deviation were recorded. \fig{actuator_load_currents_and_pos} shows the real-time current and angular position for three applied loads, calculated as the product of the lever arm and the weight: \SI{13.5}{\newton\meter}, \SI{33.5}{\newton\meter}, and \SI{44.1}{\newton\meter}. The load was attached at $t = \SI{8}{\second}$, starting with the leaver horizontally, after which the current increased accordingly. For the lowest load, the current stabilized around \SI{2}{\ampere} without a noticeable deviation in position. As the load increased, higher peak currents were observed. The current stabilizes after \SI{60}{\second}, but we observed negligible angular displacements, with a maximum of approximately 0.05 revolutions even at the highest load. The relationship between applied load and average current is summarized in \fig{actuator_current_vs_load}. The maximum load tested was \SI{73}{\newton\meter}, which required an average of \SI{8}{\ampere}, still significantly below the motor’s rated maximum.

\begin{figure}[t]
  \centering
  \includegraphics[width=\linewidth]{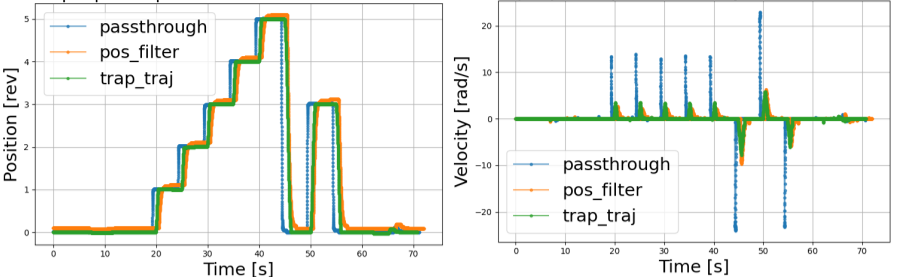}
  \caption{Position control step input response test for different input control modes. Position (left) and velocity (right) results.}
  \label{step_response_input_modes}
\end{figure}

\subsubsection{Position control step response test} This experiment aimed to compare the performance of the \textit{input modes} available in the ODrive firmware for position control. The tested modes were \textit{passthrough}, \textit{position filter}, and \textit{trapezoidal trajectory}. The \textit{inactive} mode was excluded due to excessive speed and current spikes during initial trials. \fig{step_response_input_modes} shows that all modes were capable of reaching the target position; however, \textit{passthrough} generated the highest velocity spikes. The \textit{position filter} and \textit{trapezoidal trajectory} modes produced similar velocity profiles, but the latter resulted in smoother transitions between setpoints. Based on these results, the \textit{trapezoidal trajectory} mode was selected for subsequent limb operation.

\subsection{Limb Modules}
\begin{figure}[t]
  \centering
  \includegraphics[width=\linewidth]{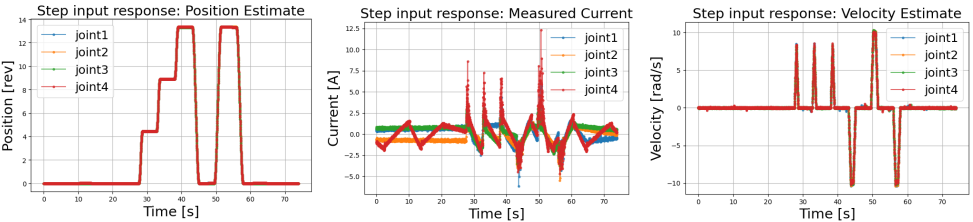}
  \caption{Step input response of all actuators in a limb module under simultaneous joint commands.}
  \label{dynamic_load_joints_response}
\end{figure}

\begin{figure}[t]
  \centering
  \includegraphics[width=\linewidth]{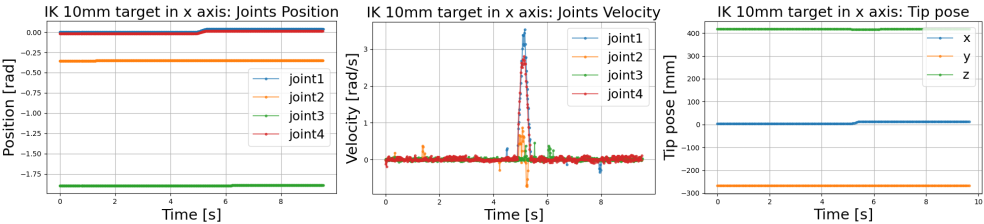}\vspace{-2mm}
  \caption{Step response Test. Actuator position control response for different input control modes.}
  \vspace{-2mm}
  \label{dynamic_load_ik}
\end{figure}

To validate actuator performance under realistic multi-joint operation, the trapezoidal trajectory mode was tested on fully assembled limb modules. \fig{dynamic_load_joints_response} presents the joint currents, positions, and speeds for simultaneous joint motions. The same target values were sent to all joints and reached without loss of synchronization. Slightly higher current peaks were observed on Joint~4, likely due to the asymmetry of the gripper assembly.
Finally, the IK mode was tested by sending Cartesian pose targets along individual axes. Particularly, a \SI{10}{mm} target on the x-axis is shown in \fig{dynamic_load_ik} as a representative case. The limb was positioned away from singular configurations before testing. All pose targets were achieved within \SI{0.5}{\second}, with smooth velocity transitions and no observable overshoot.

\section{Robot Reconfiguration Experiments}\label{experiments}
This section presents experiments demonstrating the ability of the developed modules to form a variety of robot configurations. For the current stage of work, the focus is on hardware validation and basic control software; reconfigurations were therefore performed via keyboard teleoperation or by executing pre-defined joint trajectories and IK commands. Each subsection describes the sequence used for reconfiguration, as well as the purpose and advantages of the resulting configuration.

\subsection{Limb to Limb Connection} 
\begin{figure*}[t]
  \centering
  \includegraphics[width=\linewidth]{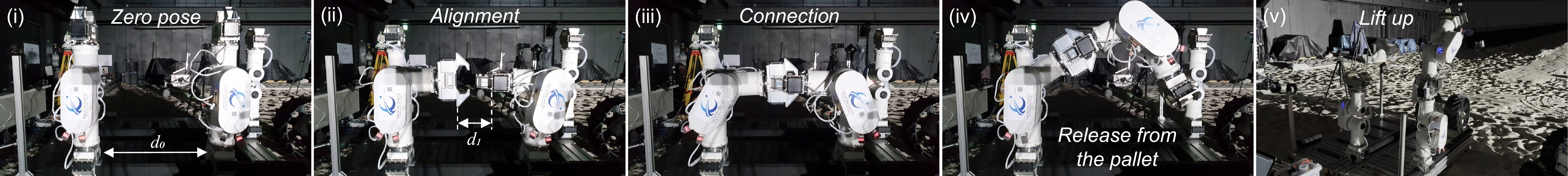}\vspace{-1mm}
  \caption{Limb to limb connection sequence.}
  \label{handshake1}
\end{figure*}
\begin{figure*}[t]
  \centering
  \includegraphics[width=\linewidth]{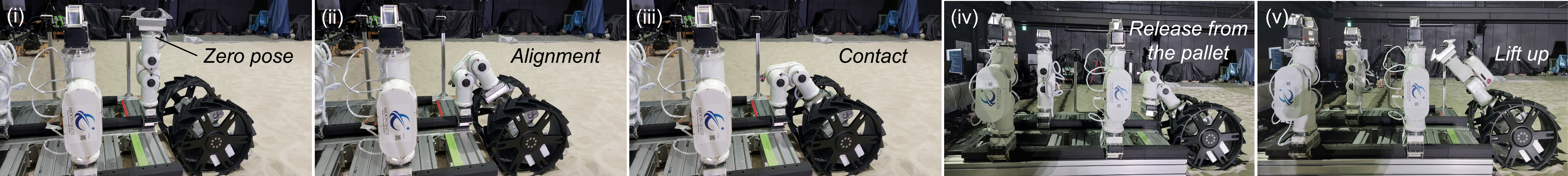}\vspace{-1mm}
  \caption{Limb to wheel connection sequence to form the \textit{Minimal} configuration.}
  \label{limb_to_wheel}
\end{figure*}
The objective of this experiment is to demonstrate how two 4-DOF limbs can be connected to form a larger 8-DOF manipulator. While a single 4-DOF limb is lighter and more energy-efficient, the combined 8-DOF arm provides redundancy and enables more versatile manipulation tasks. The connection is carried out sequentially using predefined joint angles and IK motions. As shown in \fig{handshake1}, both limbs start in the zero pose (i). They then move synchronously to a pre-connection pose (ii), chosen to facilitate alignment. Using IK, both end-effectors move horizontally towards each other by a distance $d_{1}$, determined from the initial spacing $d_{0}$. Once the grippers make contact and are aligned, they close to establish a rigid connection (iii). The lower gripper of one limb is then released from the pallet (iv), allowing the other limb to lift it (v).


\subsection{Limb to Wheel Connection} 
In this experiment, a 4-DOF limb is connected from the pallet to a single wheel module, as shown in \fig{limb_to_wheel}. The wheel modules, developed by our partner company Hero Lab and previously introduced in \cite{moonbot0}, allow the formation of mobile configurations. While a one limb plus one dual wheel combination offers limited stability, mobility is significantly improved after adding a second dual wheel module (see Section~\ref{vehicle_subsec}). The limb starts in the zero pose, positioned next to a wheel module with its orientation fixed. It then moves to a pre-connection position near the wheel’s grasping fixture using joint-angle control, followed by an IK-based trajectory to move along the fixture. After grasping the fixture, the gripper holding the limb to the pallet is opened, completing the connection.

\subsection{Vehicle Configuration}\label{vehicle_subsec} 
\begin{figure}[t]
  \centering
  \includegraphics[width=\linewidth]{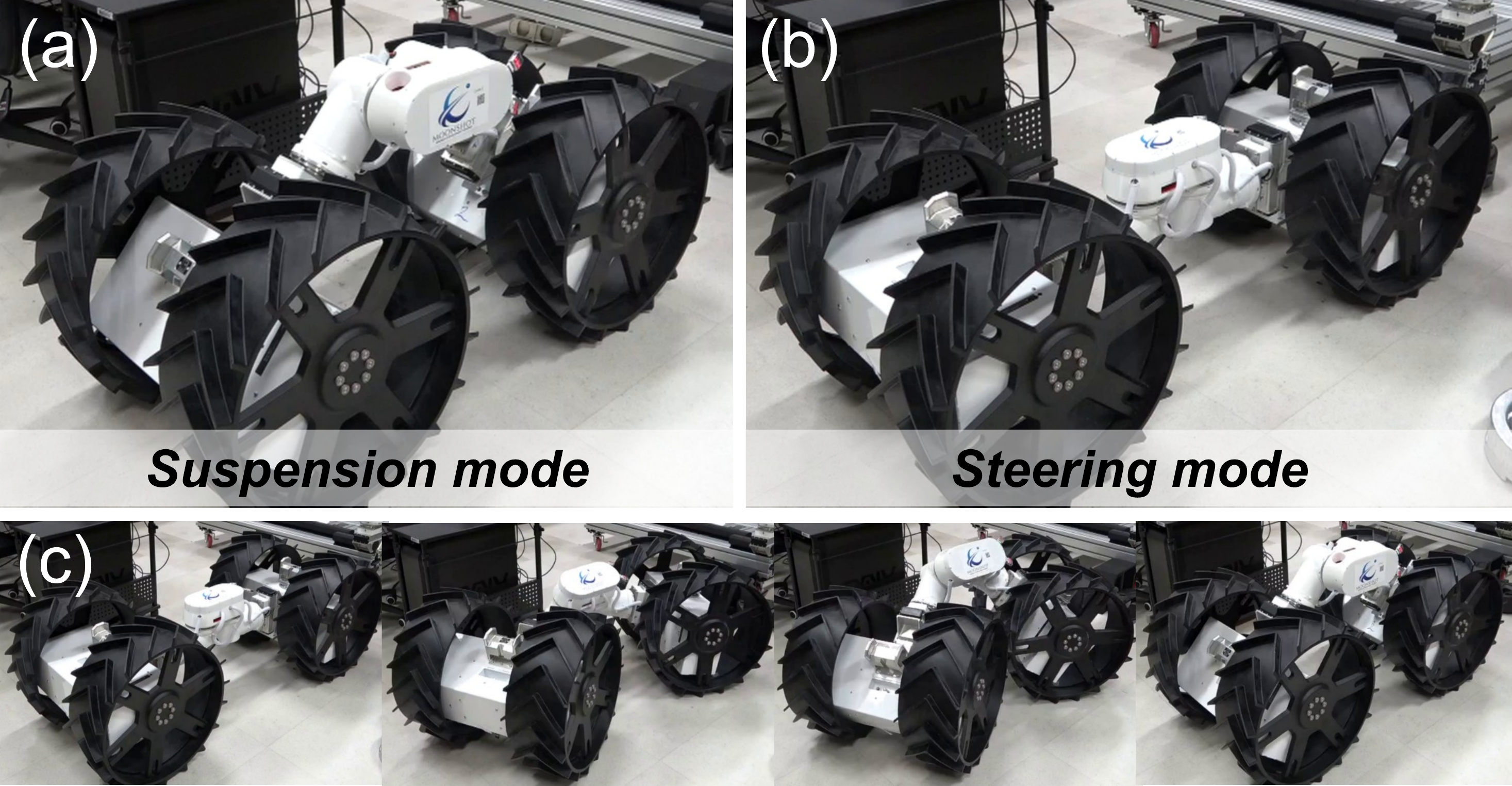}
  \caption{\emph{Vehicle} configuration formed with the developed 4-DOF Limb module. (a) Suspension mode, (b) Steering mode, and (c) Transformation sequence between modes.}
  \vspace{-2mm}
  \label{vehicle_modes}
\end{figure}

\emph{Vehicle} configuration consists of two dual wheel modules connected by one limb, as shown in \fig{vehicle_modes}. Two operational modes are possible: 
\textit{Vehicle suspension mode} (a), in which the limb’s middle pitch joints adjust wheel spacing and provide active suspension, and \textit{Vehicle steering mode} (b), in which they control steering. Transitions between these modes can be made without disconnecting the modules, as illustrated in \fig{vehicle_modes} (c).

\subsection{Dragon Configuration} 
\emph{Dragon} configuration is formed by attaching a 4-DOF limb from the pallet to an already assembled vehicle configuration (see \fig{vehicle_modes}). The connection steps are similar to the limb-to-wheel procedure but with adjusted approach angles. Once attached, the robot is stable enough to move away from the pallet and perform mobile manipulation tasks.

\begin{figure}[t]
  \centering
  \includegraphics[width=.65\linewidth]{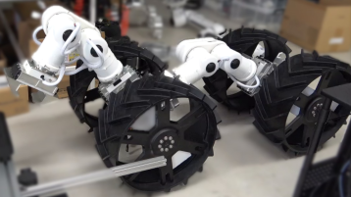}
  \caption{\emph{Dragon} configuration after connecting one limb from the palette to the vehicle configuration.}
  \vspace{-2mm}
  \label{mini_dragon}
\end{figure}

\begin{figure}[t]
  \centering
  \includegraphics[width=\linewidth]{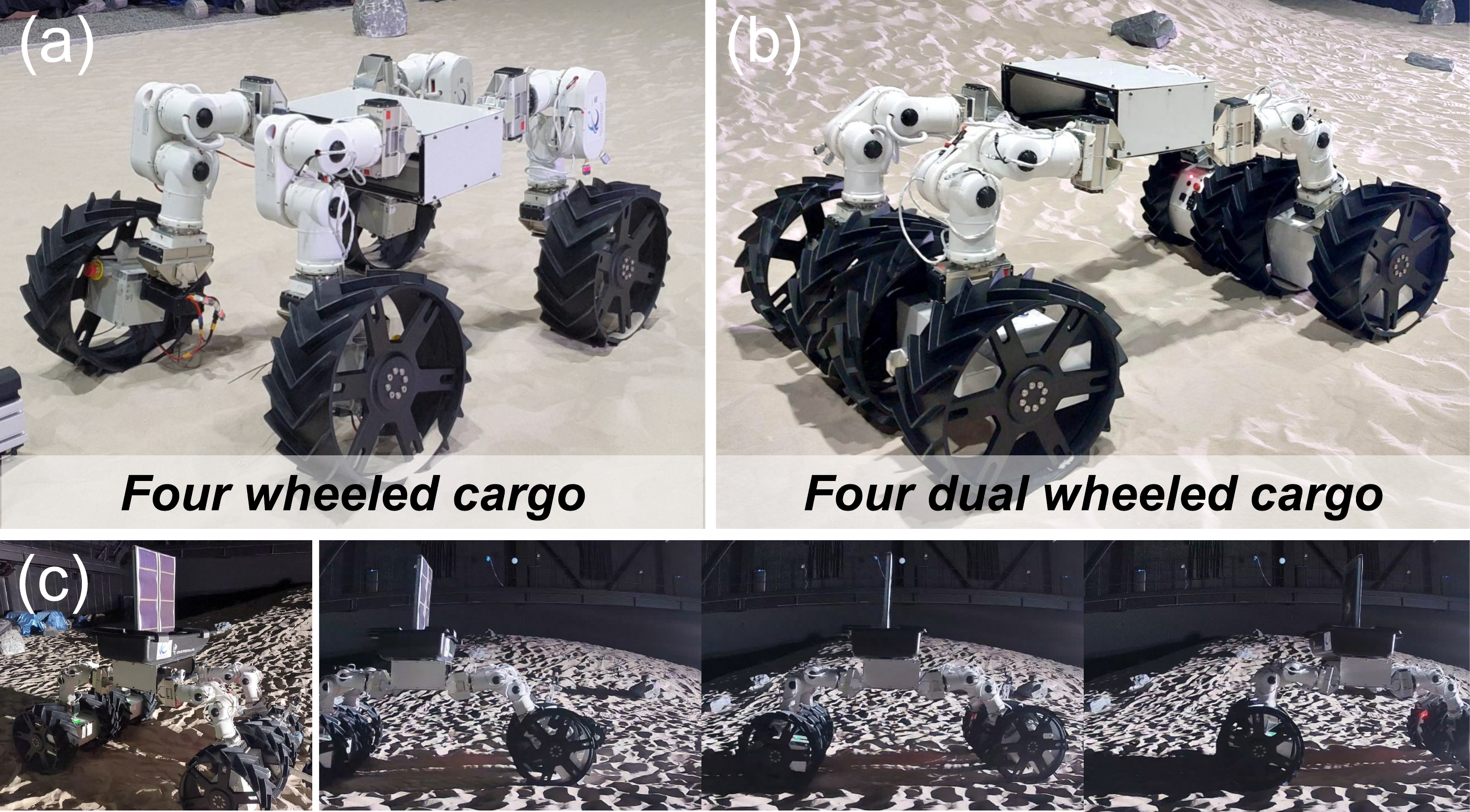}
  \caption{\emph{Cargo} configurations with four limbs plus (a) four single wheel modules and (b) four dual-wheel modules. (c) Dual-wheeled cargo configuration stably carried solar panel station mock-up on the sand.}
  \vspace{-2mm}
  \label{cargo}
\end{figure}


\subsection{Cargo Configuration} 
\emph{Cargo} configuration consists of a central body, four limbs, and four wheel modules. It provides increased payload capacity compared to Vehicle modes. We tested two types of wheel modules (single wheel and dual wheel) for Cargo configuration (see \fig{cargo}). Four wheeled Cargo enables small-scale transport tasks on the top platform, and the other Cargo with four dual-wheel modules enables larger payload transport compared to another. In our tests, the robot carried a sled with a solar panel mock-up. The maximum linear speed achievable with the dual-wheel modules is up to approximately \SI{1.0}{m/s}.



\subsection{Other Configurations} 
\begin{figure}[t!]
  \centering
  \includegraphics[width=\linewidth]{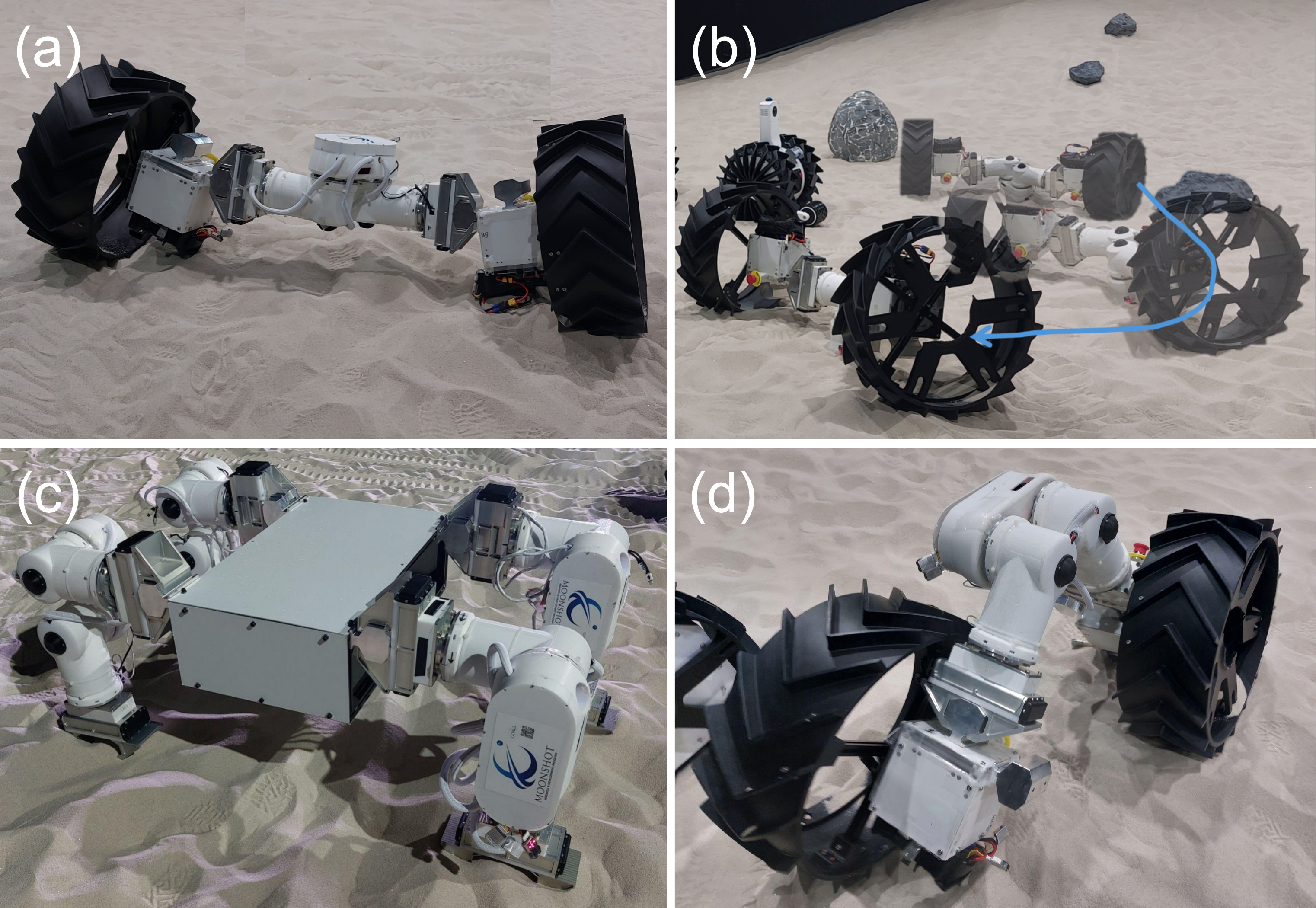}\vspace{-1mm}
  \caption{Experimental robot configurations. (a) \emph{Spinbot} configuration, (b) \emph{Spinbot} traversing the sand field, (c) \emph{Quadruped} configuration. (d) \emph{Bike} configuration.}
  \vspace*{-4mm}
  \label{otherconfigs}
\end{figure}

Additional experimental configurations include: 
\emph{Quadruped} mode (see \fig{otherconfigs}(c)), offering improved rough-terrain traversal compared to wheeled configurations; \emph{Spinbot} mode (a--b), comprising one limb and two single-wheel modules, which moves by shifting its center of gravity. This mode is designed to operate with limited resources and particularly in areas with space constraints. We were able to traverse \SI{4}{\meter} in \SI{8}{\minute} with this mode; and \emph{Bike} mode (d), also consisting of two single-wheel modules and one limb, but using the middle pitch joints for suspension, this mode is designed for narrow-passage navigation, such as between rocks. The bike mode requires an inertial measurement unit (IMU)-based balance controller, which is under development.

\section{Conclusion}\label{conclusions}
This work presented the design, development, and evaluation of modular robotic limbs and wheel modules intended for versatile reconfiguration in planetary exploration and construction tasks. The proposed system consists of 4-DOF limbs and wheel assemblies that share a common actuator, simplifying maintenance and improving interchangeability between modules. The actuator’s torque and speed characteristics were experimentally evaluated, confirming its suitability for both locomotion and manipulation in various configurations.
Nine distinct modes were demonstrated, ranging from single-limb to complex multi-limb configurations such as quadrupeds and hybrid wheeled-legged systems. This modularity allows the robot to adapt to different tasks and terrains, making it suitable also for terrestrial applications in unstructured environments.
We emphasized compactness, mechanical robustness, and software flexibility, enabling rapid integration into new arrangements without significant hardware modifications.
Future work will focus on improving autonomous reconfiguration capabilities, enhancing sensing for navigation and manipulation. These will contribute to the deployment of adaptable robotic systems capable of supporting long-term planetary construction and exploration missions.


\section*{Acknowledgement}
The authors also deeply thank Prof. Shigeo Hirose and Dr. Naoto Kimura and all the members in HERO Lab., Hakusan Corp., for their invaluable supports in the preliminary design of the modular robots.
\bibliography{./IEEEabrv,reference}

\end{document}

%% file: macros.tex
\newcommand{\fig}[1]{Fig.~\ref{#1}}

\newcommand{\tab}[1]{Table~\ref{#1}}

\def\epsgaiji#1{\leavevmode\kern-0.025zw\raise-.37zh\hbox{%
  \epsfile{file=#1,width=1.05zw}}\kern-0.025zw}
\newcommand{\MARU}[1]{{\ooalign{\hfil#1\/\hfil\crcr\raise.167ex\hbox{\mathhexbox20D}}}}

\usepackage{array}
